\documentclass[letterpaper]{article}
\usepackage{aaai}
\usepackage{times}
\usepackage{helvet}
\usepackage{courier}

\usepackage{listings}
\usepackage{algorithm}
\usepackage[noend]{algpseudocode}

\usepackage{amsthm}
\usepackage{amsmath}
\usepackage{amsfonts}
\usepackage{nicefrac} 
\usepackage{cases}
\usepackage{bm}
\usepackage{mathtools}

\usepackage{subcaption}

\newtheorem*{remark}{Remark}

\usepackage{tikz}
\usepackage{wrapfig}
\usetikzlibrary{external,positioning}
\usepackage{pgfplots}
\pgfplotsset{compat=1.14, legend style={font=\tiny},label style={font=\small}, tick label style={font=\small},}
\usetikzlibrary{matrix}
\usepgfplotslibrary{fillbetween, groupplots}

\newcommand*{\pca}{\text{PCA}}
\newcommand*{\era}{\text{ERA}}

\newtheorem{theorem}{Theorem}
\newtheorem{definition}{Definition}
\newtheorem{lemma}{Lemma}

\DeclareMathOperator*{\argmax}{\arg{}\,\max}

\newcommand*{\vp}{\mathbf{p}}
\newcommand*{\vt}{\mathbf{t}}
\newcommand*{\vx}{\mathbf{x}}

\newcommand*{\vb}{\mathbf{b}}

\pgfplotsset{
  /pgfplots/line legend with two lines/.style 2 args={
    legend image code/.code={
      \draw[mark phase=2, mark repeat=2, #1] plot coordinates {(0cm, -0.1cm) (0.25cm, -0.1cm) (0.5cm, -0.1cm)};
      \draw[mark phase=2, mark repeat=2, #2] plot coordinates {(0cm, 0.1cm) (0.25cm, 0.1cm) (0.5cm, 0.1cm)};
    }
  }
}

\definecolor{codegreen}{rgb}{0,0.6,0}
\definecolor{codegray}{rgb}{0.5,0.5,0.5}
\definecolor{codepurple}{rgb}{0.58,0,0.82}
\definecolor{backcolour}{rgb}{0.95,0.95,0.92}
 
\lstdefinestyle{mystyle}{
    backgroundcolor=\color{white},   
    commentstyle=\color{cycle6},
    keywordstyle=\color{cycle3},
    stringstyle=\color{cycle1},
    basicstyle=\ttfamily\footnotesize,
    breakatwhitespace=false,         
    breaklines=false,                 
    captionpos=b,                    
    keepspaces=true,                 
    numbers=none,                    
    numbersep=5pt,                  
    showspaces=false,                
    showstringspaces=false,
    showtabs=false,                  
    tabsize=2
}
 
\definecolor{cadmiumgreen}{rgb}{0.0, 0.42, 0.24}

\definecolor{cycle1}{RGB}{64, 83, 211}
\definecolor{cycle2}{RGB}{221, 179, 16}
\definecolor{cycle3}{RGB}{181, 29, 20}
\definecolor{cycle4}{RGB}{0, 190, 255}
\definecolor{cycle5}{RGB}{251, 73, 176}
\definecolor{cycle6}{RGB}{0, 178, 93}
\definecolor{cyclegray}{RGB}{202, 202, 202}

\usepackage{placeins}

\usepackage{multirow}

\title{CC-Cert: A Probabilistic Approach to Certify General Robustness of Neural Networks}

\author {
    Mikhail Pautov\textsuperscript{\rm 1},
    Nurislam Tursynbek\textsuperscript{\rm 1},
    Marina Munkhoeva\textsuperscript{\rm 1}, \AND
    Nikita Muravev\textsuperscript{\rm 2,3},
    Aleksandr Petiushko\textsuperscript{\rm 2,3,4},
    Ivan Oseledets\textsuperscript{\rm 1} \AND \\
    \textsuperscript{\rm 1}Skolkovo Institute of Science and Technology,
    \textsuperscript{\rm 2}Lomonosov MSU,
    \textsuperscript{\rm 3}Huawei Moscow Research Center,
    \textsuperscript{\rm 4}AIRI, Moscow\\
    mikhail.pautov@skoltech.ru, nurislam.tursynbek@gmail.com, marina.munkhoeva@skolkovotech.ru,\\ muravev.nikita@huawei.com, petyushko@yandex.ru, I.Oseledets@skoltech.ru
}

\begin{document}
\maketitle
\begin{abstract}
In safety-critical machine learning applications, it is  crucial to defend models against adversarial attacks --- small modifications of the input that change the predictions. Besides rigorously studied $\ell_p$-bounded additive perturbations, semantic perturbations (e.g. rotation, translation) raise a serious concern on deploying ML systems in real-world. Therefore, it is important to provide provable guarantees for deep learning models against semantically meaningful input transformations. In this paper, we propose a new universal probabilistic certification approach based on Chernoff-Cramer bounds that can be used in general attack settings. We estimate the probability of a model to fail if the attack is sampled from a certain distribution. Our theoretical findings are supported by experimental results on different datasets.
\end{abstract}

\section{Introduction}
Deep neural network (DNN) models have achieved tremendous success in many tasks.
On the other hand, it is well known that they are intriguingly susceptible to adversarial attacks of different kinds \cite{szegedy2013intriguing}, thus there is a pressing lack of models that are robust to such attacks. Several mechanisms are proposed as empirical defenses to various known adversarial perturbations. However, these defenses have been later were circumvented by new more aggressive attacks \cite{carlini2017towards,carlini2017adversarial,athalye2018obfuscated,tramer2020adaptive}. 

Therefore, a natural question arises: given a DNN model $f$, can we provide any provable guarantee on its prediction under a certain threat model, i.e. $f(x) = f(x_T)$, where $x_T$ is a transformed input? This is precisely the topic of a growing field of \emph{certified robustness}. A part of the existing research relies on the analysis of the Lipschitz properties of a classifier: its output's Lipschitz-continuity leads to a robustness certificate for additive attacks \cite{anil2019sorting,li2019preventing,serrurier2021achieving}.  Another excellent tool for such a task is \emph{smoothing:} the inference of the model is replaced by  averaging of predictions over the set of transformed inputs. First, this approach was developed for small-norm attacks \cite{cohen2019certified,levine2020robustness}, but then  it was successfully generalized to a much more general class of transforms such as translations or rotations \cite{balunovic2019certifying,fischer2020certified}. A recent work \cite{li2021tss} and its follow-up \cite{alfarra2021deformrs} study a general smoothing framework that is able to provide useful certificates for the most common synthetic image modifications and their compositions.



However, a drawback of smoothing methods is that they can only certify a smoothed model which is significantly slower than the original one due to a large number of samples to approximate the  expectation. Thus, we need a method that a) can certify any given black-box model, and b) is general and not tailored to a specific threat model (such as small-norm perturbations). Providing an exact rigorous certification in this setting is a very challenging task.

Instead, in this paper, we propose and investigate a probabilistic approach that produces robustness guarantees of a black-box model against different threat models. In this approach, we estimate the probability of misclassification, if the attack is sampled randomly from the admissible set of attacks. We bound this quantity by the probability of the large deviation of a certain random variable $Z$, which is derived from the distance between the probability vectors (see Lemma 1). To estimate the probability of the large deviation, we propose to use the empirical version of the Chernoff-Cramer bound and show that under the additional assumption on $Z$ this bound holds with high probability (see Theorem 1). 

\textbf{Our contributions are summarized as follows:}
\begin{itemize}
    \item We propose a new framework called Chernoff-Cramer Certification (CC-Cert) for probabilistic robustness bounds and theoretically justify them based on the empirical version of Chernoff-Cramer inequality;
    \item We test those bounds and demonstrate their efficacy for several models trained in different ways for single semantic parametric transformations and some of their compositions.
\end{itemize}

\section{Preliminaries}


\begin{figure*}[ht!]
\begin{centering}
\includegraphics[width=1\textwidth]{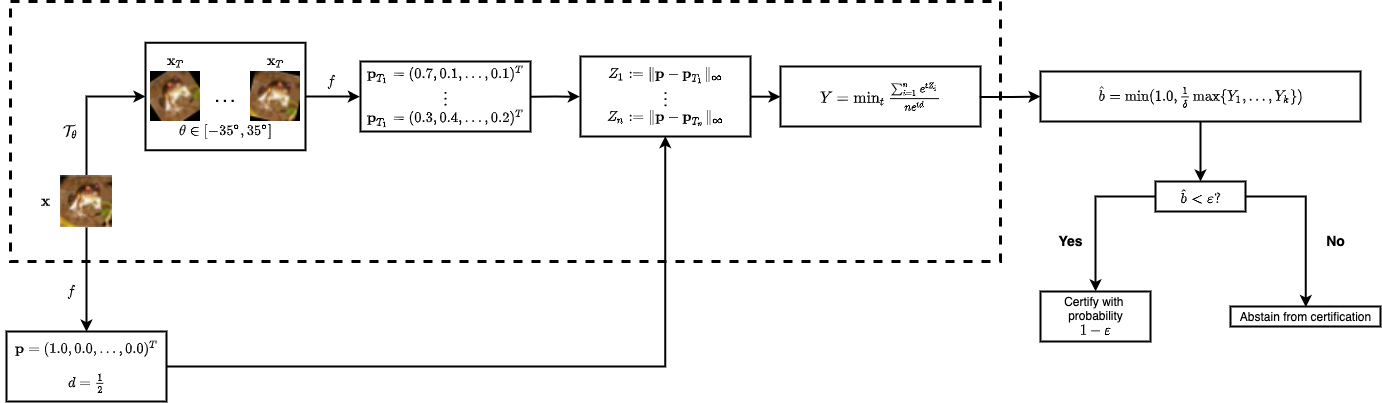}
\caption{Illustration of Algorithm \eqref{alg:bound} for a single sample. Classifier $f$ is used to compute vector $\vp$ of class probabilities and the difference $d$ between its two largest components for initial sample $x$. Then, initial sample is perturbed by transformation $\mathcal{T}_\theta$  to obtain its $n$ perturbed versions $\{x_{T_i}\}_{i=1}^n$. Classifier $f$ is then applied to these perturbed samples to obtain corresponding probability vectors $\{\vp_{T_i}\}_{i=1}^n$ and discrepancies  $\{Z_i\}_{i=1}^n$. Next, given the discrepancies $\{Z_i\}_{i=1}^n$ and set $\vt$ of temperatures, an empirical Chernoff bound is compute. The procedure of computation of bound is repeated $k$ times and resulting maximum bound is divided by $\delta$ in order to minimize the probability of bound obtained to be less than the actual one in Chernoff-Cramer method. Finally, the obtained quantity $\hat{b}$ is treated as an upper bound for the probability of change of classifier's prediction  under transform $\mathcal{T}_\theta$ of its input $x$ which is  certified with probability $1 - \varepsilon$ if $\hat{b} < \varepsilon$.} 
\label{fig:bound_compute_visual}
\end{centering}
\end{figure*}

We consider the classification task with fixed set of data samples ${\mathcal{S} = \{(x_1, y_1), \dots, (x_m, y_m)\}}$ where $x_i\in\mathbb{R}^D$ is a $D-$dimensional input drawn from unknown data distribution $p_\text{data}$ and $y_i \in \{1, \dots, K\}$ are corresponding labels. Let ${f: \mathbb{R}^D \to [0,1]^{K}}$ be the deterministic function such that ${ \displaystyle h(x) = \argmax_{k \in \{1, \dots, K\}} f_k(x)}$ maps any input $x$ to the corresponding data label. 

Under a transformation ${T: \mathbb{R}^D \to \mathbb{R}^D}$, an input image $x$ is transformed into $x_T=T(x)$. It is known that $f(x)$ and $f(x_T)$ may differ significantly under certain transformations, namely, some perturbations which do not change semantic of an image may mislead the classifier $h(\cdot)$ that correctly classifies its unperturbed version.  In our work, we focus on certifying the stability of $h(\cdot)$ under transformation~$T$ with certain assumptions on its parameter space~$\Theta$, or, in other words, providing guarantees that $h(\cdot)$ is \textit{probabilistically robust} at $x$ under perturbation of general type. 
A more formal description of such perturbations is given in
\begin{definition}
Perturbation of general type is a parametric mapping $T:  \Theta \times \mathcal{X} \to \mathcal{X}.$ Throughout the paper, we denote $\Theta$ as a parameter space of a given transform, $T_\theta(x) = x_T$ as the transformed version of $x$ given $\theta \in \Theta$ and $\mathbb{S}_T(x)$ as the space of all transformed versions of $x$ under perturbation~$T.$

\end{definition}

\begin{definition}[Probabilistic robustness, \cite{mohapatra2020higher}]
Let  $T$ be the transformation of general type and $x$ be the input sample with ground truth class $c$. Suppose that $\mathbb{S}_T(x)$ is the space of all images  $x_T$ of $x$ under perturbations induced by $T$ and $\mathbb{P}$ be the probability measure on space $
\mathbb{S}_T(x)$.
The $K$-class model $f$ is said to be robust at point $x$ with probability at least $1-\epsilon$ if   
\begin{equation}
    \displaystyle \mathbb{P}_{x_T \sim \mathbb{S}_T(x)} \left( \argmax_{k \in \{1, \dots, K\}} f_k(x_T) = c \right) \ge 1-\epsilon.
\label{eq:prob_robustness}
\end{equation}
\label{def:prob_robustness}
\end{definition}


\section{Probabilistic Certification of Robustness}

\subsection{Robustness and Large Deviations}
Our proposal is to certify the robustness of the classifier $f$ by estimating the probability of large deviations of its output probability vectors. The following lemma shows how the discrepancy $f(x) - f(x_T)$ between the output vectors $f(x)$ and $f(x_T)$ assigned by $f$ to the original sample $x$ and its transformed version $x_T$ is connected to their predicted class labels.

\begin{lemma}
Let $f: \mathbb{R}^D \to [0,1]^C$ be the classifier with a known inference procedure and $x$ be an input sample. Let $x_T$ be a transformation of $x$ after transform $T \in \mathcal{T}$ where $\mathcal{T}$ is a given class of transforms.
Assume that ${\vp=f(x)}$ and ${\vp_T=f(x_T)}$ are the probability vectors of the original sample $x$ and its transformed version $x_T$, respectively. Let ${c=\arg\max \vp}$ and ${\tilde{c} = \arg\max \vp_T}$ be the classes assigned by $f$ to $x$ and $x_T$, respectively and let  $d=\frac{p_1-p_2}{2}$ be the half of the difference between two largest components of~$\vp.$  

Then, if $\|\vp - \vp_T\|_\infty<d$ holds, $\tilde{c} = c$.
\end{lemma}
\begin{proof}
Suppose that ${c \ne \tilde{c}}$, then, ${{\vp_T}_{\tilde{c}} > {\vp_T}_c}$ and ${\vp_c > \vp_{\tilde{c}}.}$ On the other hand, since the difference norm ${\|\vp - \vp_T\|_\infty = \max(|\vp_1 - \vp_{T_1}|, \dots, |\vp_C - {\vp_T}_C|) < d,}$ we have $$\vp_c - d < {\vp_T}_c\ \text{and}\ {\vp_T}_{\tilde{c}} - d < \vp_{\tilde{c}},$$ which leads to $${\vp_T}_c - {\vp_T}_{\tilde{c}} > (\vp_c - d) - (\vp_{\tilde{c}} + d) = \vp_c - \vp_{\tilde{c}} - 2d.$$
Since $2d = p_1-p_2 = \vp_c - p_2,$ $$\vp_c - \vp_{\tilde{c}} - 2d = p_2 - \vp_{\tilde{c}} \ge 0,$$ yielding a contradiction. Thus, $c=\tilde{c}.$
\end{proof}

Intuitively, the lemma states that whenever the maximum change in the output $\vp_T$ does not exceed half the distance $d$ between the two largest components in $\vp$, the argument of the maximum in $\vp_T$ does not change in comparison to the one in $\vp$.

We suggest to estimate the right tail of the probability distribution of $Z = Z_{x, \mathcal{T}} = \|\vp-\vp_T\|_\infty$. Namely, we want to bound from above the probability $\mathbb{P}\left(Z \ge d\right)$ to provide guarantees that transformations from class $\mathcal{T}$ applied to $x$ do not lead to a change of class label assigned by $f.$

\subsection{Deriving the Bound: Estimates of the Tail of the Distribution}

In this section, we describe our method to estimate the tail of probability distribution. Namely, we refer to the Chernoff-Cramer approach and propose a technique that allows us to use a sample mean instead of the population mean in the right-hand side of the bound.

\begin{lemma}[Markov's inequality]
Let $Z$ be a non-negative scalar random variable with finite expectation and $t \in \mathbb{R}, t>0$, then

$$\mathbb{P} (Z \ge t) \le \frac{\mathbb{E} (Z)}{t}.$$
\end{lemma}

\begin{lemma}[Chernoff-Cramer method, \cite{boucheron2003concentration}]

Let $Z$ be a scalar random variable, $t \in \mathbb{R}, t>0$ and $d \in \mathbb{R}$. Then 
\begin{equation}\label{eq:bern}
\mathbb{P} (Z \ge d) = \mathbb{P} (e^{Zt} \ge e^{dt}) \le \frac{\mathbb{E} (e^{Zt})}{e^{dt}}
\end{equation}
 by a Markov's inequality. 

\end{lemma}
The right-hand side of Eq. \ref{eq:bern} is the upper-bound for $\epsilon$ in Eq. \ref{eq:prob_robustness}. Note that depending on the value of $t$, Eq. \ref{eq:bern} produces a lot of bounds, so it is natural to choose $t$ that minimizes its right-hand side.

\subsubsection{Sample Mean Instead of True Expectation.} Overall, we want to use Eq.~\eqref{eq:bern} to bound the probability that the deviation between $\vp$ and $\vp_T$ is greater than half the difference between the two largest components of $\vp$, i.e. $d$. Unfortunately, it is impractical to compute its right hand side  directly, because the population mean of $e^{Zt}$ is unknown in general. Instead, the task is to estimate the density of 
\begin{equation}
    Y=
    \exp(-dt)
    \frac1n\sum_{i=1}^n\exp(Z_it),
    \label{eq:main}
\end{equation}
where $\forall i \in \{1,\dots n\}$ $Z_i = \|\vp-\vp_{T_i}\|_\infty$ is the norm of the difference in the probability vectors of original and transformed samples. 
There is a certain challenge in such an approach: when the population mean is replaced by the sample mean, it is possible to underestimate the true expectation, and thus, provide an incorrect bound which is less than the right-hand side of Eq.~\eqref{eq:bern}. It implies that the inequality 
\begin{equation}\label{chern_with_sample}
    \mathbb{P}(Z \ge d) \le Y,
\end{equation} which is the modification of \eqref{eq:bern} by the replacement of the true expectation by a sample mean, holds with some probability: it is \textit{guaranteed to hold} only in case  ${Y \ge e^{-dt}\mathbb{E}(e^{Zt})}.$

Thus, the probability of the sample mean to underestimate the population mean,
\begin{equation}\label{eq:samplevspop}
\mathbb{P} 
\left(Y \le  \frac{\mathbb{E}(e^{Zt})}{e^{dt}}\right), 
\end{equation}
needs to be small as it regulates the correctness of computation of the bound in the form of Eq.~\eqref{eq:main}. Our proposal is to bound from above the probability in  Eq.~\eqref{eq:samplevspop} by sampling $k$ i.i.d. sample means $\{Y_1, \dots, Y_k\}$ in the form of \eqref{eq:main} and exploit the statistics of the random variable $\max(\{Y_1, \dots, Y_k\}).$ 

The pseudo-code in Algorithm~\ref{alg:bound} summarizes the bound computation for a single input $\vx$\footnote{In our repository the code is parallelized for batches.}. 

\begin{figure}[H]
\begin{algorithm}[H]
    \begin{algorithmic}
    \Function{bound}{$f, \vx, y, \vt, \texttt{transform}, n, k, \delta$}
        \State{$\vp \gets f(\vx)$}
    	\Comment{compute model output on original $x$} {}
    	\State{$d = \frac{\vp[0] - \vp[1]}{2}$}
    	\Comment{Compute the difference between top 2 classes}{}
	   	\State{$\hat{y} = \texttt{max}(\vp)$}
	   	\Comment{compute predicted class} {}
		\State{$\texttt{hit} \gets \hat{y} == y$}
		\Comment{boolean variable indicating correctness of prediction for initial input} {}
    	\State{$\vx_{n} = \texttt{repeat}(\vx, n)$}
    	\Comment{repeat original input $n$ times}
    	\State{$\vp_{n} = \texttt{repeat}(\vp, n)$}
    	\Comment{repeat $\vp$ $n$ times}
    	\State{$\texttt{bounds} \gets \{0, \dots, 0\}$}
    	\Comment{initialize placeholder for bounds with $k$ zeros}
    	\For{\texttt{i=0, i<k, i++}}
        \State{$\vx_T = \texttt{transform}(\vx_{n})$}
    	\Comment{apply random transforms to input $\vx_{n}$} {}
        \State{$\vp_T \gets f(\vx_T)$}
    	\Comment{compute model output on transformed $\vx_T$} {}
    	\State{$\textbf{Z} = \| \vp_{n} - \vp_T \|_\infty$}
    	\Comment{compute the change in output vector in terms of max norm} {}
    	\State{$\textbf{Z} \vt^\top = \texttt{outer}(\textbf{Z}, \vt)$}
    	\Comment{multiply every diff by vector of temperatures $\vt$}
    	\State{$\mathbb{E}(e^{\textbf{Z}\vt}) = \texttt{mean}_{0}(\texttt{exp}(\textbf{Z} \vt^\top))$}
    	\Comment{compute sample mean over $n$ random transforms}
    	\State{$e^{d\textbf{t}} = \texttt{exp}(d \vt)$}
    	\State{$\vb = \mathbb{E}(e^{\textbf{Z}\vt}) / e^{d\vt}$}
    	\Comment{get the bounds vector over temperatures}
    	\State{$b_\text{min} = \texttt{min} (\vb)$} 
    	\Comment{compute the smallest bound over temperatures} {}
    	\State{$\texttt{bounds}[i] \gets b_\text{min}$}
    	\EndFor
    	\State{$\texttt{bound} = \frac{1}{\delta}\texttt{max}( \texttt{bounds})$}
    	\Comment{compute the largest of obtained bounds and normalize by $\delta$}
        \State{\Return{$\texttt{min}(1.0, \texttt{bound})$}, \texttt{hit}}
    \EndFunction{}
    \end{algorithmic}
    \caption{Chernoff-Cramer bound calculation algorithm.}
    \label{alg:bound}
\end{algorithm}
\end{figure}




\subsubsection{Several Sample Means Instead of One.}
In this section, we show that, under certain assumptions, bound provided by the Algorithm~\eqref{alg:bound} may be used instead of the right-hand side of Eq.~\eqref{eq:bern}.

\begin{lemma}[Paley-Zygmund, \cite{paley1930some}] Suppose a random variable $X$ is positive and have finite variance, $\sigma^2_X < \infty$. Then, $\forall \delta \in (0,1)$,  

\begin{equation}\label{paley-zygmund}
\mathbb{P}\left(X < \delta \mathbb{E}(X)\right) \le \frac{\sigma^2_X}{\sigma^2_X + (1-\delta)^2 (\mathbb{E}(X))^2}.
\end{equation}
\end{lemma}

\begin{theorem}[Worst-out-of-k-bounds]
Suppose that random variable $X$ takes values from $[0,1]$, probability density function of random variable $\xi=e^{Xt}$ is positively skewed and has
coefficient of variation $C_v=\frac{\sigma_\xi}{\mathbb{E}(\xi)} \sim 1$. Then, given $\hat{b}$ as the bound produced by the Algorithm~\ref{alg:bound}, 

\begin{equation}\label{power}
    \mathbb{P}\left(\hat{b} < \frac{\mathbb{E}(\xi)}{e^{dt}}\right) < \left(\frac{1}{1 + \frac{{n}(1-\delta)^2}{C^2_v}}\right)^k.
\end{equation}
\end{theorem}

\begin{figure}[ht!]
\begin{centering}
\includegraphics[width=0.45\textwidth]{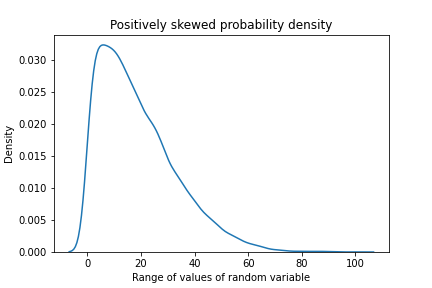}
\caption{Example of positively skewed probability density function.} 
\label{fig:pos_skew}
\end{centering}
\end{figure}

\begin{proof}
As one of the steps of the Algorithm~ \ref{alg:bound}, sample means $Y_j= \overline{\xi}_je^{-dt} = \left\{ \frac{\sum_{i=1}^{{n}} e^{tX_i}}{{n} e^{dt}}\right\}_j$ are computed. Note that the expectation and variance of $\xi$ and $\overline{\xi}$ are related as  ${\mathbb{E}(\overline{\xi}) = \mathbb{E}(\xi) = \mu_\xi}$ and $\sigma^2_{\overline{\xi}} = \frac{\sigma^2_\xi}{{n}}$, respectively. 

Then, according to \eqref{paley-zygmund},

\begin{align*}
\begin{split}
\mathbb{P}\left(\overline{\xi} < \delta  \mathbb{E}(\overline{\xi})\right) &= \mathbb{P}\left(\overline{\xi}  < \delta  \mu_\xi\right) \le 
 \frac{ \sigma^2_{\overline{\xi}}}{ \sigma^2_{\overline{\xi}} + (1-\delta)^2(\mu_\xi)^2}  \\
 &= \frac{ \sigma^2_\xi}{ \sigma^2_\xi + {n}(1-\delta)^2\mu^2_\xi} =  
 \frac{1}{1 + \frac{ {n}(1-\delta)^2}{C^2_v}} \\ 
 &= p({n}, C_v). 
 \end{split}
 \end{align*}
 
 Since sample means $\overline{\xi}_j = Y_j e^{dt}, j \in [1,\dots,k],$ in Algorithm \eqref{alg:bound} are i.i.d., 
 
 \begin{equation*}
 \begin{aligned}
     \mathbb{P}\left(\max(\overline{\xi_1}, \dots, \overline{\xi_k}\right) < \delta \mu_{\xi}) &= \prod_{j=1}^k \mathbb{P}\left(\xi_j < \delta \mu_{\xi}\right) \\&\le \prod_{j=1}^k p({n}, C_v) = p({n}, C_v)^k.
\end{aligned}
 \end{equation*}
 
 Thus, since the left-hand side of the above inequality is the one from \eqref{power}, namely
 
  \begin{equation*}
  \begin{aligned}
  \mathbb{P}\left(\max(\overline{\xi_1}, \dots, \overline{\xi_k}) < \delta \mu_{\xi}\right) &= \mathbb{P}\left(\frac{1}{\delta}\max(\overline{\xi_1}, \dots, \overline{\xi_k}) < \mu_{\xi}\right) \\ &= \mathbb{P}\left(\hat{b} < \frac{\mathbb{E}(\xi)}{e^{dt}}\right),
\end{aligned}
 \end{equation*}
 
 inequality \eqref{power} holds.
\end{proof}
\begin{remark}
When the number of samples in the Algorithm ~\ref{alg:bound} is enough, namely, $n>n_0 = (1-\delta)^{-2}C^2_v,$ the right hand side of inequality \eqref{power} is less than $\left(\frac{1}{2}\right)^k$. In other words, there exists a number $n_0$ such that for all $n > n_0$ the probability can be made  small for not very large $k$. In Figure \ref{fig:pos_skew}, we present an example of positively skewed probability density.
\end{remark}

\subsection{Types of Model Training and Computation Cost of Inference}

It should be mentioned that our method can be applied for certification of any model $f(x)$, i.e. it does not require changing the inference procedure as it is done in smoothing methods. However, if the model training is done in the standard way, the overall robust accuracy is expected to be significantly lower  than the plain accuracy of the model. A straightforward way to improve the robustness of the model is to modify the training procedure. We can adapt the training procedure used in smoothing methods such as \cite{li2021tss}. Such procedure can be implemented through data augmentation: the model is trained on transformed samples, thus we expect it will have higher robust accuracy. During the inference stage, the smoothed model has to  be evaluated many times \cite{horvath2021boosting}, thus increasing the complexity. In our approach, we can apply the certification directly to the original model trained in an \emph{augmented} way. Our numerical experiments confirm that such models typically have better accuracy, however, there are notable exceptions. This means that more research is needed in training more robust plain models and/or mixed approaches such as smoothing with a small number of samples.
\section{Experiments}
In order to evaluate our method, we assess the proposed bounds on the public datasets, namely, MNIST and CIFAR-10. 

All the models, training and testing procedures are implemented in \texttt{PyTorch} \cite{paszke2019pytorch}, and the transformations considered are implemented with use of $\texttt{Kornia}$ framework \cite{riba2020kornia}. We use architectures, training parameters and procedures from \cite{li2021tss} and evaluate our approach on $500$ random test samples for each experiment, following \cite{cohen2019certified}.

As a result of the experiments, we provide probabilistically certified accuracy, \textit{\pca}, in dependence on probability threshold $\varepsilon$ and show how it is connected to empirical robust accuracy, \textit{\era}, under corresponding adaptive attack. Namely, given the classifier $h(\cdot)$, set of images ${\mathcal{S} = \{(x_1, y_1), \dots, (x_m, y_m)\}}$ and threshold $\varepsilon$, probabilistically certified accuracy is computed as

\begin{equation*}\displaystyle
    \pca(\mathcal{S}, \varepsilon) = \frac{\left|  (x, y) \in \mathcal{S}: \texttt{BOUND}(x) < \varepsilon \ \&\ h(x)=y \right|}{m}.
\end{equation*}

At the same time, given the discretization  ${\Theta = \{\theta_1, \dots, \theta_r\}}$  of space of parameters of the transform $T$, empirical robust accuracy is computed as a fraction of objects from $\mathcal{S}$ that are correctly classified under all the transformations $T_{\theta_i}, i \in [1, \dots, r]:$

\begin{equation*}\displaystyle
    \era(\mathcal{S}) = \frac{\left|  (x, y) \in \mathcal{S}: h(T_{\theta_i}(x)) = y \ \forall i \in [1, \dots r] \right|}{m}.
\end{equation*}



\subsection{Bound Computation Parameters and Cost}
In all our experiments, we use the following parameters for the Algorithm~\ref{alg:bound}: number of samples ${n=200}$, number of bounds ${k=30}$, parameter ${\delta=0.9}$ for Lemma (4), temperature vector $\vt$ consists of $500$ evenly spaced numbers from interval $[10^{-4}, 10^{4}]$. We use $r=20$ as discretization parameter for space of parameters for each transform. Execution of the Algorithm~\ref{alg:bound} on single GPU Tesla V100-SXM2-16GB takes ${30-1000}$ seconds on MNIST and ${60-2000}$ seconds on CIFAR-10 depending on the transform complexity.

\subsection{Considered Transformations} Here we present the set of transformations studied in this work and provide the set of corresponding parameters used in the experiments with single transformations.

\paragraph{Image Rotation.} The transformation of rotation is parameterized by the rotation angle $\phi$ and is assumed to be followed by the bilinear interpolation with zero-padding of rotated images.
In case of CIFAR-10 dataset, we use ${\Theta = \{\phi \in [-10^{\circ}, 10^{\circ}]\}}$, and  ${\Theta = \{\phi \in [-50^{\circ}, 50^{\circ}]\}}$ for MNIST.
\paragraph{Image Translation.} The transformation of image translation  is parameterized by a  translation vector and is also  assumed  to be followed by bilinear interpolation. We use zero-padding and are not restricted to discrete translations. In case of CIFAR-10, we use $\Theta = \{v \in \mathbb{R}^2: |v| \le 0.2 * w\}$, and for MNIST  $\Theta = \{v \in \mathbb{R}^2: |v| \le 0.3 * w\}$, with $w$ equal to the width/height of the image, i.e. translation by no more than $20\%$ of image for CIFAR-10 and no more than $30\%$ of image for MNIST, respectively.

\begin{figure}[ht!]
\begin{centering}
\includegraphics[width=0.45\textwidth]{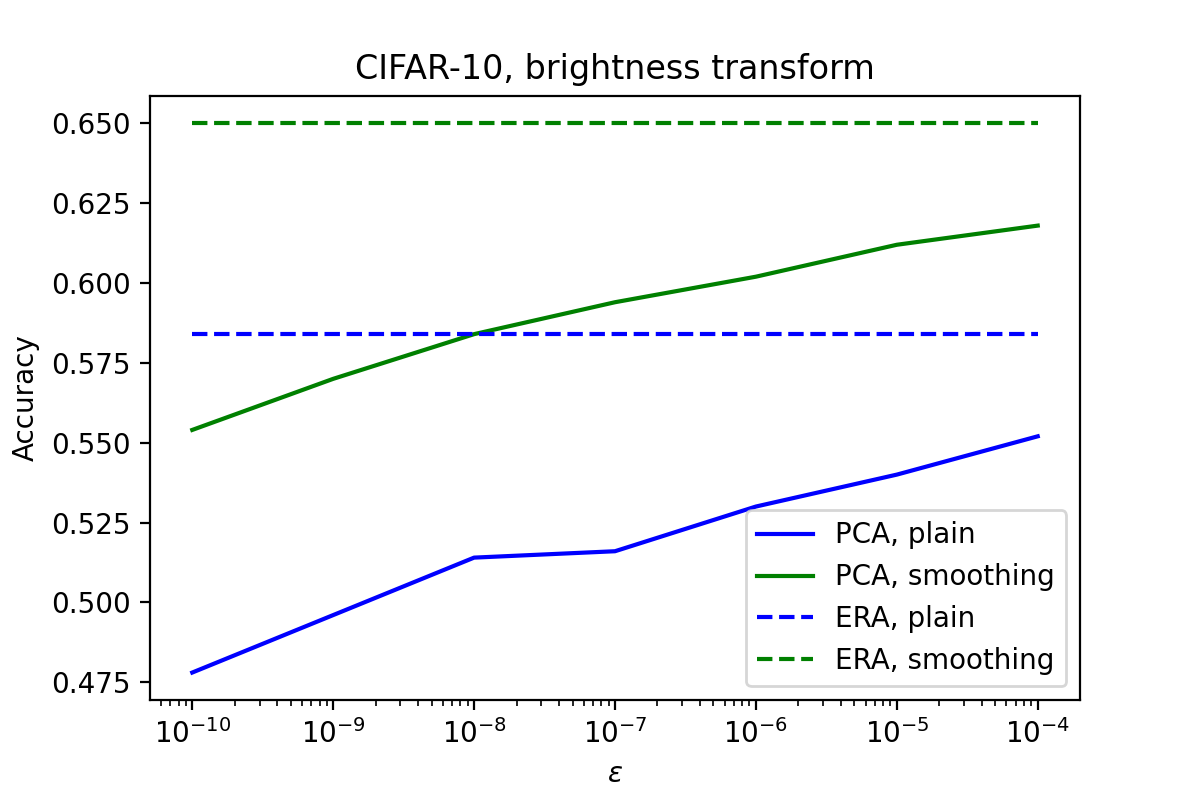}
\caption{An illustration of the evaluation protocol on CIFAR-10 dataset, brightness adjustment transform. Number of samples ${n=20000}$, number of computed bounds ${k=30}$, ${\delta=0.9}$, brightness adjustment parameter ${\theta_b \in [-40\%, +40\%].}$} 
\label{fig:cifar10_brighntess}
\end{centering}
\end{figure}

\paragraph{Brightness Adjustment.}
The transformation of brightness adjustment is just an element-wise addition of a  brightness factor to an image. We change the brightness by no more than $40\%$ for CIFAR-10 and by no more than $50\%$ for MNIST. 

\paragraph{Contrast Adjustment.}
The transformation of contrast adjustment is just an element-wise multiplication of image by a contrast factor. We change contrast by no more than $40\%$ for CIFAR-10 and by no more than $50\%$ for MNIST.  

\paragraph{Image Scaling.}
The transform of image scaling is resize of one by two factors corresponding to spatial dimensions. In our work, we zero-pad  downscaled images and use ${\Theta = \{s \in \mathbb{R}^2: \|s\|_2^2 \in [0.7, 1.3]\}}$ for CIFAR-10 and ${\Theta = \{s \in \mathbb{R}^2: \|s\|_2^2 \in [0.5, 1.5]\}}$ for MNIST. In other words, we adjust no more than $30\%$ of image size on CIFAR-10 and no more than $50\%$ of image size on MNIST.

\paragraph{Gaussian Blurring.}
The transformation of Gaussian blurring \cite{li2021tss} convolves an image with the Gaussian function $$G_\sigma(k) = \frac{1}{\sqrt{2 \pi \sigma}} e^{-k^2 / 2\sigma},$$ where a squared kernel radius $\sigma$ is a parameter. In our experiments, ${\Theta = \{\sigma \in (0, 9]\}}$ for both CIFAR-10 and MNIST.  

\paragraph{Composition of Transformations.} 
Aside from single transformations, we consider more complicated image perturbations, namely compositions of transformations. For example, the composition of translation and rotation of an image is a consecutive application of (a) rotation and (b) translation, with two steps of interpolation en route.

\paragraph{Quantitative Results of Experiments.}
We present  considered transformations, corresponding parameters, and quantitative results in Table \ref{table_res} and visualize our results in the Technical Appendix. In almost all cases, PCA bounds are smaller than ERA, and the difference is small. This confirms that our bounds are sufficiently tight for a wide class of transformations.

\begin{table*}[ht!]
\centering
\caption{Comparison of probabilistically certified accuracy and empirical robust accuracy. We report probabilistically certified accuracy for three levels of threshold parameter $\varepsilon$: high confidence in certification ($\varepsilon < 10^{-10}$), middle level of confidence ($\varepsilon < 10^{-7}$) and low level of confidence ($\varepsilon < 10^{-4}$). In the column \textbf{PA}, we report initial accuracy on the whole datasets.}\smallskip
\scalebox{0.85}{
\begin{tabular}{|c|c|c|c|c|c|c|c|c|}\hline
\multirow{2}{*}{\textbf{Dataset}} & \multirow{2}{*}{\textbf{Transform}} & \multirow{2}{*}{\textbf{Parameters}} & \multirow{2}{*}{\textbf{Training type}}&  \multirow{2}{*}{\textbf{ERA}} &  \multicolumn{3}{c|}{\textbf{PCA($\varepsilon$)}} & \multirow{2}{*}{\textbf{PA}}\\
\cline{6-8}
& & & & &$\varepsilon = 10^{-10}$& $\varepsilon = 10^{-7}$ & $\varepsilon = 10^{-4}$ & \\
\hline
\hline
\multirow{18}{*}{CIFAR-10} &  \multirow{2}{*}{Brightness} & \multirow{2}{*}{$\theta_b \in [-40\%, 40\%]$} & plain  & $58.4\%$ & $47.8\%$ & $51.6\%$ & $55.2\%$ & $91.18\%$\\
\cline{4-9}
 & & & smoothing  &$65.0\%$ & $55.4\%$ & $59.4\%$ & $61.8\%$ & $88.67\%$\\
\cline{2-9}
& \multirow{2}{*}{Contrast}  & \multirow{2}{*}{$\theta_c \in [-40\%, 40\%]$} & plain& $91.6\%$ & $62.4\%$ & $67.0\%$ & $69.6\%$ & $91.18\%$\\
\cline{4-9}
 & & & smoothing & $88.0\%$ & $67.0\%$ & $72.8\%$ & $74.2\%$ & $88.67\%$\\
\cline{2-9}
 & \multirow{2}{*}{Rotation} & \multirow{2}{*}{$\theta_r \in [-10^{\circ}, 10^{\circ}]$} & plain & $73.4\%$ & $64.6\%$ & $69.0\%$ & $71.0\%$ & $91.18\%$\\
\cline{4-9}
& & & smoothing &$72.4\%$ & $57.4\%$ & $63.6\%$ & $67.4\%$ & $87.77\%$\\
\cline{2-9}
 & \multirow{2}{*}{Gaussian blur} & \multirow{2}{*}{$\theta_g \in [0, 3]$ -- kernel radius} & plain & $12.2\%$ & $11.0\%$ & $11.0\%$ & $11.0\%$ & $91.18\%$\\
\cline{4-9}
&   &  & smoothing & $60.4\%$ & $57.2\%$ & $57.2\%$ & $57.8\%$ & $81.11\%$\\
\cline{2-9}
 & \multirow{2}{*}{Translation} & \multirow{2}{*}{$|\theta_t| \le 20\%$}& plain & $40.4\%$ & $28.0\%$ & $31.2\%$ & $35.2\%$ & $91.18\%$\\
\cline{4-9}
 &   &  & smoothing & $35.0\%$ & $17.8\%$ & $22.4\%$ & $25.6\%$ & $85.98\%$\\
\cline{2-9}
 & \multirow{2}{*}{Scale} & \multirow{2}{*}{$\theta_s \in [70\%, 130\%]$} & plain & $57.0\%$ & $54.4\%$ & $54.4\%$ & $54.4\%$ & $91.18\%$\\
\cline{4-9}
 &   &  & smoothing &$55.0\%$ & $53.4\%$ & $53.4\%$ & $53.6\%$ & $86.76\%$\\
\cline{2-9}
 & \multirow{2}{*}{Contrast + Brightness}  & \multirow{2}{*}{see Contrast \& Brightness}  & plain & $0.0\%$ & $0.0\%$ & $0.0\%$ & $0.0\%$ & $91.18\%$\\
\cline{4-9}
 &   &  & smoothing & $0.4\%$ & $0.0\%$ & $0.0\%$ & $0.0\%$ & $88.67\%$\\
\cline{2-9}
& \multirow{2}{*}{Rotation + Brightness} &  \multirow{2}{*}{see Rotation \& Brightness} & plain & $22.6\%$ & $16.2\%$ & $20.6\%$ & $21.8\%$ & $91.18\%$\\
\cline{4-9}
 &  &  & smoothing & $30.4\%$ & $21.2\%$ & $24.6\%$ & $27.6\%$ & $84.50\%$\\
\cline{2-9}
 & \multirow{2}{*}{Scale + Brightness} & \multirow{2}{*}{see Scale \& Brightness} & plain & $10.2\%$ & $10.4\%$ & $10.4\%$ & $10.4\%$ & $91.18\%$\\
\cline{4-9}
 &  &  & smoothing & $41.8\%$ & $40.6\%$ & $40.6\%$ & $40.6\%$ & $86.53\%$\\
\hline
\hline
\multirow{18}{*}{MNIST} & \multirow{2}{*}{Brightness}  & \multirow{2}{*}{$\theta_b \in [-50\%, 50\%]$} & plain & $97.8\%$ & $94.8\%$ & $96.4\%$ & $97.0\%$ & $99.26\%$\\
\cline{4-9}
 & & & smoothing & $98.6\%$ & $97.0\%$ & $98.2\%$ & $98.2\%$ & $99.04\%$\\
\cline{2-9}
& \multirow{2}{*}{Contrast}  & \multirow{2}{*}{$\theta_c \in [-50\%, 50\%]$} & plain& $98.8\%$ & $96.0\%$ & $97.0\%$ & $97.2\%$ & $99.26\%$\\
\cline{4-9}
 &   &  & smoothing & $98.6\%$ & $98.2\%$ & $98.2\%$ & $98.2\%$ & $99.04\%$\\
\cline{2-9}
 &  \multirow{2}{*}{Rotation} & \multirow{2}{*}{$\theta_r \in [-50^{\circ}, 50^{\circ}]$} & plain & $18.8\%$ & $11.6\%$ & $14.8\%$ & $16.4\%$ & $99.26\%$\\
\cline{4-9}
&   &  & smoothing & $98.0\%$ & $97.0\%$ & $97.4\%$ & $97.6\%$ & $99.01\%$\\
\cline{2-9}
 &  \multirow{2}{*}{Gaussian blur} & \multirow{2}{*}{$\theta_g \in [0, 3]$ -- kernel radius} & plain & $78.0\%$ & $68.8\%$ & $68.8\%$ & $68.8\%$ & $99.26\%$\\
\cline{4-9}
&   &  & smoothing & $97.8\%$ & $97.8\%$ & $97.8\%$ & $97.8\%$ & $98.35\%$\\
\cline{2-9}
 & \multirow{2}{*}{Translation} & \multirow{2}{*}{$|\theta_t| \le 30\%$} & plain & $0.0\%$ & $0.0\%$ & $0.0\%$ & $0.0\%$ & $99.26\%$\\
\cline{4-9}
 &   &  & smoothing & $39.6\%$ & $31.4\%$ & $34.4\%$ & $38.2\%$ & $99.09\%$\\
\cline{2-9}
 & \multirow{2}{*}{Scale} & \multirow{2}{*}{$\theta_s \in [70\%, 130\%]$} & plain & $21.6\%$ & $21.0\%$ & $21.0\%$ & $21.0\%$ & $99.26\%$\\
\cline{4-9}
 &   &  & smoothing & $34.4\%$ & $34.4\%$ & $34.4\%$ & $34.4\%$ & $99.25\%$\\
\cline{2-9}
 & \multirow{2}{*}{Contrast + Brightness} & \multirow{2}{*}{see Contrast \& Brightness} & plain & $8.4\%$ & $0.0\%$ & $0.0\%$ & $0.0\%$ & $99.26\%$\\
\cline{4-9}
 &   &  & smoothing & $7.6\%$ & $2.4\%$ & $2.4\%$ & $2.4\%$ & $99.04\%$\\
\cline{2-9}
& \multirow{2}{*}{Rotation + Brightness} & \multirow{2}{*}{see Rotation \& Brightness} & plain & $14.0\%$ & $9.2\%$ & $11.2\%$ & $13.0\%$ & $99.26\%$\\
\cline{4-9}
 &  &  & smoothing & $95.2\%$ & $93.0\%$ & $93.4\%$ & $94.6\%$ & $99.08\%$\\
\cline{2-9}
 & \multirow{2}{*}{Scale + Brightness} & \multirow{2}{*}{see Scale \& Brightness} & plain & $13.0\%$ & $13.4\%$ & $13.4\%$ & $13.4\%$ & $99.26\%$ \\
\cline{4-9}
 &  &  & smoothing & $93.4\%$ & $93.0\%$ & $93.0\%$ & $93.4\%$ & $99.37\%$\\
\hline
\end{tabular}}
\label{table_res}
\end{table*}

\subsection{Evaluation Protocol Illustration}

Figure \ref{fig:cifar10_brighntess} illustrates the certification of the model to brightness adjustment on CIFAR-10. Following Theorem 1, it is at least $1 - \left(\frac{1}{2}\right)^{30}$ chance that almost $60\%$ of considered samples will be correctly classified by a considered model trained with smoothing after no more than $40\%$ brightness adjustment with probability $1-\varepsilon \ge 1 - 10^{-7}.$

\subsection{Comparison with the Clopper-Pearson Confidence Intervals}

It is natural to compare upper bound for probability of a model to fail obtained via proposed method with the upper limit of corresponding Clopper-Pearson confidence interval \cite{clopper1934use} for an unknown  binomial probability $p$. Namely, we randomly sample $n$ transforms and count the number $k$ of the ones that mislead a classifier. If it is zero, we can upper bound the probability ${p \le \frac{C(\alpha)}{n}}$ given confidence level $\alpha$. Thus, in order to get bound of~$10^{-4}$ we need at least $10000$ transformed version of a sample. Analogously to \textit{probabilistically certified accuracy},  \textit{Clopper-Pearson certified accuracy} (CPCA) is computed as $$\text{CPCA}(\mathcal{S}, \varepsilon) = \frac{\left|  (x, y) \in \mathcal{S}: \texttt{CP}(x) < \varepsilon \ \&\ h(x)=y \right|}{m},$$ where $\texttt{CP}(x) = \texttt{CP}(x, n, k, \alpha)$ is the upper limit of the classical Clopper-Pearson bound for sample $x$ given $n,k,\alpha$ from above.

\begin{figure*}[ht!]
\centering
\begin{minipage}[b]{0.48\textwidth}
\includegraphics[width=\textwidth]{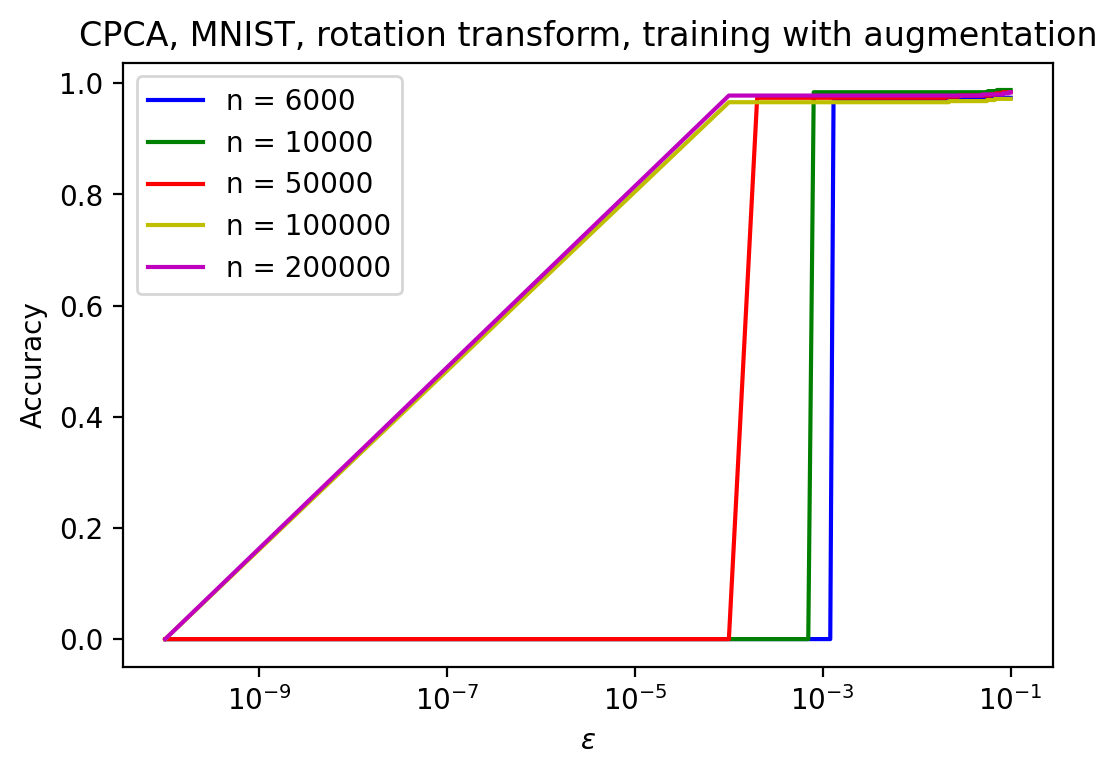}
\caption{Clopper-Pearson certified accuracy depending on number of trials $n$ for experiment with rotation transform on MNIST.} 
\label{fig:cpca}
\end{minipage}\hfill
\begin{minipage}[b]{0.48\textwidth}
\includegraphics[width=\textwidth]{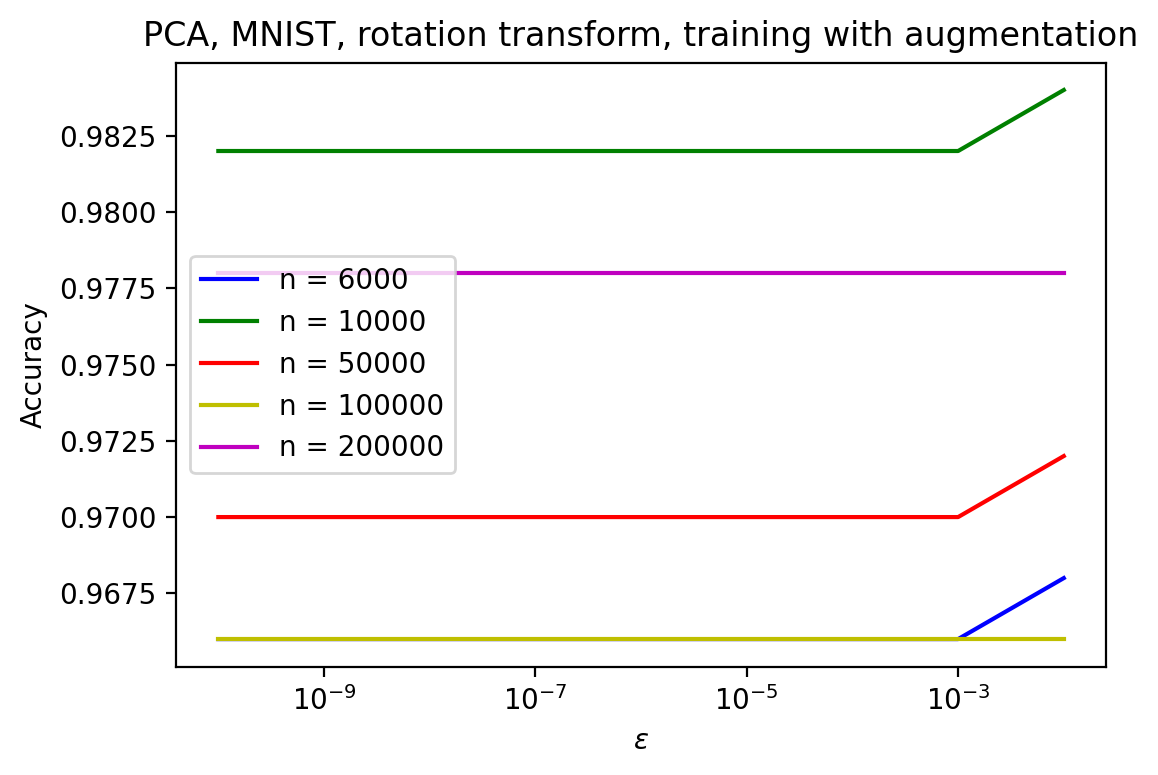}
\caption{Probabilistically certified accuracy depending on number of samples $n$ for experiment with rotation transform on MNIST.} 
\label{fig:pca}
\end{minipage}
\end{figure*}

In Figures \ref{fig:cpca},  \ref{fig:pca}, we compare Clopper-Pearson certified accuracy for different number of trials $n$ and probabilistically certified accuracy for the same experiment, respectively.

It can be seen that the method proposed in this paper allows to get a better bound with smaller number of samples. The reason is that we use additional information -- the distribution of the random variable $Z$, unlike the CP approach which relies solely on the fact of misclassification.
\section{Related Work}
\textbf{Adversarial Vulnerability.} Adversarial examples \cite{szegedy2013intriguing,goodfellow2014explaining} have attracted significant attention due to vulnerability of neural networks in safety-critical scenarios. Although initially found as $\ell_p$-bounded attacks \cite{goodfellow2014explaining,moosavi2016deepfool}, several beyond $\ell_p$ adversaries have been proposed that degrade performance of deep models. They include semantic perturbations, such as rotation and translation \cite{kanbak2018geometric,engstrom2018rotation,xiao2018spatially}, color manipulation \cite{hosseini2018semantic}, renderer-based lightness and geometry transformation \cite{liu2018beyond}, parametric attribute manipulation \cite{joshi2019semantic}, 3d scene parameters such as camera pose, sun position, global translation, rotation \cite{li2018differentiable}. These threats present a serious obstacle for model deployment in safety-critical real-world applications. 

\noindent\textbf{Certified Defenses Against Additive Perturbations.} To overcome the  adversarial vulnerability of deep models, several empirical defenses have been presented, however, they were found to fail against new adaptive attacks \cite{tramer2020adaptive}. This forced researchers to provide certified defenses that guarantee stable performance. A number of works have been tackling this problem for small additive perturbations from different perspectives. This includes semi-definite programming \cite{raghunathan2018semidefinite}, interval bound propagation \cite{gowal2018effectiveness}, convex relaxation \cite{wong2018provable}, duality perspective \cite{dvijotham2018dual}, Satisfiability Modulo Theory \cite{bunel2017unified,katz2017reluplex,ehlers2017formal}. Although these methods consider small networks, randomized smoothing \cite{cohen2019certified,li2018certified,lecuyer2019certified} and its improvements \cite{salman2019provably,zhai2020macer} have been proposed as a certified defense for any network that can be applied to large-scale datasets.

\noindent\textbf{Certified Defenses Against Image Transformations.}
Following certificates for $\ell_p$-bounded perturbations, several methods proposed certified robustness against semantic perturbations. Generating linear relaxation and propagating through a neural network was used in \cite{singh2019abstract,balunovic2019certifying,mohapatra2020towards}. Extensions of  MCMC-based randomized smoothing \cite{cohen2019certified} to the space of semantic transformations were presented in \cite{fischer2020certified,li2021tss,alfarra2021deformrs}. Our work presents a probabilistic approach for certifying neural networks against different semantic perturbations. Perhaps, the closest to our approach are PROVEN \cite{weng2019proven}, where authors present a framework to probabilistically certify neural networks against $\ell_p$ perturbations and work of \cite{mangal2019robustness}, where authors certify a neural network in a probabilistic way by overapproximation of input regions that violate model's robustness. However, these works do not consider probabilistic certifications under semantic transformations.
\section{Conclusion and Future Work}

We proposed a new framework CC-Cert for probabilistic certification of the robustness of DNN models to synthetic transformations and provide theoretical bounds on its robustness guarantees. Our experiments show the applicability of our approach for certification of model's robustness to a random transformation of input. An important question that is not considered in our work is the worst-case analysis of robustness to synthetic transformations. Future work includes analysis of the discrepancy between worst-case transformations of general type and corresponding random transformations. A possible gap between certification with probability and certification in worst-case should be carefully analyzed and narrowed.

\section{Acknowledgements}
This work was partially supported by a grant for research centers in the field of artificial intelligence, provided by the Analytical Center for the Government of the Russian Federation in accordance with the subsidy agreement (agreement identifier 000000D730321P5Q0002) and the agreement with the Ivannikov Institute for System Programming of the Russian Academy of Sciences dated November 2, 2021 No. 70-2021-00142.

\bibliography{main}

\begin{thebibliography}{}

\bibitem[\protect\citeauthoryear{Alfarra \bgroup et al\mbox.\egroup
  }{2021}]{alfarra2021deformrs}
Alfarra, M.; Bibi, A.; Khan, N.; Torr, P.~H.; and Ghanem, B.
\newblock 2021.
\newblock Deformrs: Certifying input deformations with randomized smoothing.
\newblock {\em arXiv preprint arXiv:2107.00996}.

\bibitem[\protect\citeauthoryear{Anil, Lucas, and
  Grosse}{2019}]{anil2019sorting}
Anil, C.; Lucas, J.; and Grosse, R.
\newblock 2019.
\newblock Sorting out {Lipschitz} function approximation.
\newblock In {\em International Conference on Machine Learning},  291--301.
\newblock PMLR.

\bibitem[\protect\citeauthoryear{Athalye, Carlini, and
  Wagner}{2018}]{athalye2018obfuscated}
Athalye, A.; Carlini, N.; and Wagner, D.
\newblock 2018.
\newblock Obfuscated gradients give a false sense of security: Circumventing
  defenses to adversarial examples.
\newblock In {\em International conference on machine learning},  274--283.
\newblock PMLR.

\bibitem[\protect\citeauthoryear{Balunovi{\'c} \bgroup et al\mbox.\egroup
  }{2019}]{balunovic2019certifying}
Balunovi{\'c}, M.; Baader, M.; Singh, G.; Gehr, T.; and Vechev, M.
\newblock 2019.
\newblock Certifying geometric robustness of neural networks.
\newblock {\em Advances in Neural Information Processing Systems 32}.

\bibitem[\protect\citeauthoryear{Boucheron, Lugosi, and
  Bousquet}{2003}]{boucheron2003concentration}
Boucheron, S.; Lugosi, G.; and Bousquet, O.
\newblock 2003.
\newblock Concentration inequalities.
\newblock In {\em Summer School on Machine Learning},  208--240.
\newblock Springer.

\bibitem[\protect\citeauthoryear{Bunel \bgroup et al\mbox.\egroup
  }{2017}]{bunel2017unified}
Bunel, R.; Turkaslan, I.; Torr, P.~H.; Kohli, P.; and Kumar, M.~P.
\newblock 2017.
\newblock A unified view of piecewise linear neural network verification.
\newblock {\em arXiv preprint arXiv:1711.00455}.

\bibitem[\protect\citeauthoryear{Carlini and
  Wagner}{2017a}]{carlini2017adversarial}
Carlini, N., and Wagner, D.
\newblock 2017a.
\newblock Adversarial examples are not easily detected: Bypassing ten detection
  methods.
\newblock In {\em Proceedings of the 10th ACM workshop on artificial
  intelligence and security},  3--14.

\bibitem[\protect\citeauthoryear{Carlini and
  Wagner}{2017b}]{carlini2017towards}
Carlini, N., and Wagner, D.
\newblock 2017b.
\newblock Towards evaluating the robustness of neural networks.
\newblock In {\em 2017 ieee symposium on security and privacy (sp)},  39--57.
\newblock IEEE.

\bibitem[\protect\citeauthoryear{Clopper and Pearson}{1934}]{clopper1934use}
Clopper, C.~J., and Pearson, E.~S.
\newblock 1934.
\newblock The use of confidence or fiducial limits illustrated in the case of
  the binomial.
\newblock {\em Biometrika}  404--413.

\bibitem[\protect\citeauthoryear{Cohen, Rosenfeld, and
  Kolter}{2019}]{cohen2019certified}
Cohen, J.; Rosenfeld, E.; and Kolter, Z.
\newblock 2019.
\newblock Certified adversarial robustness via randomized smoothing.
\newblock In {\em International Conference on Machine Learning},  1310--1320.
\newblock PMLR.

\bibitem[\protect\citeauthoryear{Dvijotham \bgroup et al\mbox.\egroup
  }{2018}]{dvijotham2018dual}
Dvijotham, K.; Stanforth, R.; Gowal, S.; Mann, T.~A.; and Kohli, P.
\newblock 2018.
\newblock A dual approach to scalable verification of deep networks.
\newblock In {\em UAI}, volume~1, ~3.

\bibitem[\protect\citeauthoryear{Ehlers}{2017}]{ehlers2017formal}
Ehlers, R.
\newblock 2017.
\newblock Formal verification of piece-wise linear feed-forward neural
  networks.
\newblock In {\em International Symposium on Automated Technology for
  Verification and Analysis},  269--286.
\newblock Springer.

\bibitem[\protect\citeauthoryear{Engstrom \bgroup et al\mbox.\egroup
  }{2018}]{engstrom2018rotation}
Engstrom, L.; Tran, B.; Tsipras, D.; Schmidt, L.; and Madry, A.
\newblock 2018.
\newblock A rotation and a translation suffice: Fooling cnns with simple
  transformations.

\bibitem[\protect\citeauthoryear{Fischer, Baader, and
  Vechev}{2020}]{fischer2020certified}
Fischer, M.; Baader, M.; and Vechev, M.
\newblock 2020.
\newblock Certified defense to image transformations via randomized smoothing.

\bibitem[\protect\citeauthoryear{Goodfellow, Shlens, and
  Szegedy}{2014}]{goodfellow2014explaining}
Goodfellow, I.~J.; Shlens, J.; and Szegedy, C.
\newblock 2014.
\newblock Explaining and harnessing adversarial examples.
\newblock {\em arXiv preprint arXiv:1412.6572}.

\bibitem[\protect\citeauthoryear{Gowal \bgroup et al\mbox.\egroup
  }{2018}]{gowal2018effectiveness}
Gowal, S.; Dvijotham, K.; Stanforth, R.; Bunel, R.; Qin, C.; Uesato, J.;
  Arandjelovic, R.; Mann, T.; and Kohli, P.
\newblock 2018.
\newblock On the effectiveness of interval bound propagation for training
  verifiably robust models.
\newblock {\em arXiv preprint arXiv:1810.12715}.

\bibitem[\protect\citeauthoryear{Horváth \bgroup et al\mbox.\egroup
  }{2021}]{horvath2021boosting}
Horváth, M.~Z.; Müller, M.~N.; Fischer, M.; and Vechev, M.
\newblock 2021.
\newblock Boosting randomized smoothing with variance reduced classifiers.

\bibitem[\protect\citeauthoryear{Hosseini and
  Poovendran}{2018}]{hosseini2018semantic}
Hosseini, H., and Poovendran, R.
\newblock 2018.
\newblock Semantic adversarial examples.
\newblock In {\em Proceedings of the IEEE Conference on Computer Vision and
  Pattern Recognition Workshops},  1614--1619.

\bibitem[\protect\citeauthoryear{Joshi \bgroup et al\mbox.\egroup
  }{2019}]{joshi2019semantic}
Joshi, A.; Mukherjee, A.; Sarkar, S.; and Hegde, C.
\newblock 2019.
\newblock Semantic adversarial attacks: Parametric transformations that fool
  deep classifiers.
\newblock In {\em Proceedings of the IEEE/CVF International Conference on
  Computer Vision},  4773--4783.

\bibitem[\protect\citeauthoryear{Kanbak, Moosavi-Dezfooli, and
  Frossard}{2018}]{kanbak2018geometric}
Kanbak, C.; Moosavi-Dezfooli, S.-M.; and Frossard, P.
\newblock 2018.
\newblock Geometric robustness of deep networks: analysis and improvement.
\newblock In {\em Proceedings of the IEEE Conference on Computer Vision and
  Pattern Recognition},  4441--4449.

\bibitem[\protect\citeauthoryear{Katz \bgroup et al\mbox.\egroup
  }{2017}]{katz2017reluplex}
Katz, G.; Barrett, C.; Dill, D.~L.; Julian, K.; and Kochenderfer, M.~J.
\newblock 2017.
\newblock Reluplex: An efficient smt solver for verifying deep neural networks.
\newblock In {\em International Conference on Computer Aided Verification},
  97--117.
\newblock Springer.

\bibitem[\protect\citeauthoryear{Lecuyer \bgroup et al\mbox.\egroup
  }{2019}]{lecuyer2019certified}
Lecuyer, M.; Atlidakis, V.; Geambasu, R.; Hsu, D.; and Jana, S.
\newblock 2019.
\newblock Certified robustness to adversarial examples with differential
  privacy.
\newblock In {\em 2019 IEEE Symposium on Security and Privacy (SP)},  656--672.
\newblock IEEE.

\bibitem[\protect\citeauthoryear{Levine and Feizi}{2020}]{levine2020robustness}
Levine, A., and Feizi, S.
\newblock 2020.
\newblock Robustness certificates for sparse adversarial attacks by randomized
  ablation.
\newblock In {\em Proceedings of the AAAI Conference on Artificial
  Intelligence}, volume~34,  4585--4593.

\bibitem[\protect\citeauthoryear{Li \bgroup et al\mbox.\egroup
  }{2018a}]{li2018certified}
Li, B.; Chen, C.; Wang, W.; and Carin, L.
\newblock 2018a.
\newblock Certified adversarial robustness with additive noise.
\newblock {\em arXiv preprint arXiv:1809.03113}.

\bibitem[\protect\citeauthoryear{Li \bgroup et al\mbox.\egroup
  }{2018b}]{li2018differentiable}
Li, T.-M.; Aittala, M.; Durand, F.; and Lehtinen, J.
\newblock 2018b.
\newblock Differentiable monte carlo ray tracing through edge sampling.
\newblock {\em ACM Transactions on Graphics (TOG)} 37(6):1--11.

\bibitem[\protect\citeauthoryear{Li \bgroup et al\mbox.\egroup
  }{2019}]{li2019preventing}
Li, Q.; Haque, S.; Anil, C.; Lucas, J.; Grosse, R.~B.; and Jacobsen, J.-H.
\newblock 2019.
\newblock Preventing gradient attenuation in {Lipschitz} constrained
  convolutional networks.
\newblock {\em Advances in Neural Information Processing Systems}
  32:15390--15402.

\bibitem[\protect\citeauthoryear{Li \bgroup et al\mbox.\egroup
  }{2021}]{li2021tss}
Li, L.; Weber, M.; Xu, X.; Rimanic, L.; Kailkhura, B.; Xie, T.; Zhang, C.; and
  Li, B.
\newblock 2021.
\newblock {TSS}: Transformation-specific smoothing for robustness
  certification.

\bibitem[\protect\citeauthoryear{Liu \bgroup et al\mbox.\egroup
  }{2018}]{liu2018beyond}
Liu, H.-T.~D.; Tao, M.; Li, C.-L.; Nowrouzezahrai, D.; and Jacobson, A.
\newblock 2018.
\newblock Beyond pixel norm-balls: Parametric adversaries using an analytically
  differentiable renderer.
\newblock {\em arXiv preprint arXiv:1808.02651}.

\bibitem[\protect\citeauthoryear{Mangal, Nori, and
  Orso}{2019}]{mangal2019robustness}
Mangal, R.; Nori, A.~V.; and Orso, A.
\newblock 2019.
\newblock Robustness of neural networks: A probabilistic and practical
  approach.
\newblock In {\em 2019 IEEE/ACM 41st International Conference on Software
  Engineering: New Ideas and Emerging Results (ICSE-NIER)},  93--96.
\newblock IEEE.

\bibitem[\protect\citeauthoryear{Mohapatra \bgroup et al\mbox.\egroup
  }{2020a}]{mohapatra2020higher}
Mohapatra, J.; Ko, C.-Y.; Weng, T.-W.; Chen, P.-Y.; Liu, S.; and Daniel, L.
\newblock 2020a.
\newblock Higher-order certification for randomized smoothing.
\newblock {\em arXiv preprint arXiv:2010.06651}.

\bibitem[\protect\citeauthoryear{Mohapatra \bgroup et al\mbox.\egroup
  }{2020b}]{mohapatra2020towards}
Mohapatra, J.; Weng, T.-W.; Chen, P.-Y.; Liu, S.; and Daniel, L.
\newblock 2020b.
\newblock Towards verifying robustness of neural networks against a family of
  semantic perturbations.
\newblock In {\em Proceedings of the IEEE/CVF Conference on Computer Vision and
  Pattern Recognition},  244--252.

\bibitem[\protect\citeauthoryear{Moosavi-Dezfooli, Fawzi, and
  Frossard}{2016}]{moosavi2016deepfool}
Moosavi-Dezfooli, S.-M.; Fawzi, A.; and Frossard, P.
\newblock 2016.
\newblock Deepfool: a simple and accurate method to fool deep neural networks.
\newblock In {\em Proceedings of the IEEE Conference on Computer Vision and
  Pattern Recognition},  2574--2582.

\bibitem[\protect\citeauthoryear{Paley and Zygmund}{1930}]{paley1930some}
Paley, R., and Zygmund, A.
\newblock 1930.
\newblock On some series of functions,(1).
\newblock In {\em Mathematical Proceedings of the Cambridge Philosophical
  Society}, volume~26,  337--357.
\newblock Cambridge University Press.

\bibitem[\protect\citeauthoryear{Paszke \bgroup et al\mbox.\egroup
  }{2019}]{paszke2019pytorch}
Paszke, A.; Gross, S.; Massa, F.; Lerer, A.; Bradbury, J.; Chanan, G.; Killeen,
  T.; Lin, Z.; Gimelshein, N.; Antiga, L.; et~al.
\newblock 2019.
\newblock Pytorch: An imperative style, high-performance deep learning library.
\newblock {\em Advances in Neural Information Processing Systems}
  32:8026--8037.

\bibitem[\protect\citeauthoryear{Raghunathan, Steinhardt, and
  Liang}{2018}]{raghunathan2018semidefinite}
Raghunathan, A.; Steinhardt, J.; and Liang, P.
\newblock 2018.
\newblock Semidefinite relaxations for certifying robustness to adversarial
  examples.
\newblock {\em arXiv preprint arXiv:1811.01057}.

\bibitem[\protect\citeauthoryear{Riba \bgroup et al\mbox.\egroup
  }{2020}]{riba2020kornia}
Riba, E.; Mishkin, D.; Ponsa, D.; Rublee, E.; and Bradski, G.
\newblock 2020.
\newblock Kornia: an open source differentiable computer vision library for
  pytorch.
\newblock In {\em Proceedings of the IEEE/CVF Winter Conference on Applications
  of Computer Vision},  3674--3683.

\bibitem[\protect\citeauthoryear{Salman \bgroup et al\mbox.\egroup
  }{2019}]{salman2019provably}
Salman, H.; Yang, G.; Li, J.; Zhang, P.; Zhang, H.; Razenshteyn, I.; and
  Bubeck, S.
\newblock 2019.
\newblock Provably robust deep learning via adversarially trained smoothed
  classifiers.
\newblock {\em arXiv preprint arXiv:1906.04584}.

\bibitem[\protect\citeauthoryear{Serrurier \bgroup et al\mbox.\egroup
  }{2021}]{serrurier2021achieving}
Serrurier, M.; Mamalet, F.; Gonz{\'a}lez-Sanz, A.; Boissin, T.; Loubes, J.-M.;
  and del Barrio, E.
\newblock 2021.
\newblock Achieving robustness in classification using optimal transport with
  hinge regularization.
\newblock In {\em Proceedings of the IEEE/CVF Conference on Computer Vision and
  Pattern Recognition},  505--514.

\bibitem[\protect\citeauthoryear{Singh \bgroup et al\mbox.\egroup
  }{2019}]{singh2019abstract}
Singh, G.; Gehr, T.; P{\"u}schel, M.; and Vechev, M.
\newblock 2019.
\newblock An abstract domain for certifying neural networks.
\newblock {\em Proceedings of the ACM on Programming Languages} 3(POPL):1--30.

\bibitem[\protect\citeauthoryear{Szegedy \bgroup et al\mbox.\egroup
  }{2013}]{szegedy2013intriguing}
Szegedy, C.; Zaremba, W.; Sutskever, I.; Bruna, J.; Erhan, D.; Goodfellow, I.;
  and Fergus, R.
\newblock 2013.
\newblock Intriguing properties of neural networks.
\newblock {\em arXiv preprint arXiv:1312.6199}.

\bibitem[\protect\citeauthoryear{Tramer \bgroup et al\mbox.\egroup
  }{2020}]{tramer2020adaptive}
Tramer, F.; Carlini, N.; Brendel, W.; and Madry, A.
\newblock 2020.
\newblock On adaptive attacks to adversarial example defenses.
\newblock {\em arXiv preprint arXiv:2002.08347}.

\bibitem[\protect\citeauthoryear{Weng \bgroup et al\mbox.\egroup
  }{2019}]{weng2019proven}
Weng, L.; Chen, P.-Y.; Nguyen, L.; Squillante, M.; Boopathy, A.; Oseledets, I.;
  and Daniel, L.
\newblock 2019.
\newblock {PROVEN}: Verifying robustness of neural networks with a
  probabilistic approach.
\newblock In {\em International Conference on Machine Learning},  6727--6736.
\newblock PMLR.

\bibitem[\protect\citeauthoryear{Wong and Kolter}{2018}]{wong2018provable}
Wong, E., and Kolter, Z.
\newblock 2018.
\newblock Provable defenses against adversarial examples via the convex outer
  adversarial polytope.
\newblock In {\em International Conference on Machine Learning},  5286--5295.
\newblock PMLR.

\bibitem[\protect\citeauthoryear{Xiao \bgroup et al\mbox.\egroup
  }{2018}]{xiao2018spatially}
Xiao, C.; Zhu, J.-Y.; Li, B.; He, W.; Liu, M.; and Song, D.
\newblock 2018.
\newblock Spatially transformed adversarial examples.
\newblock {\em arXiv preprint arXiv:1801.02612}.

\bibitem[\protect\citeauthoryear{Zhai \bgroup et al\mbox.\egroup
  }{2020}]{zhai2020macer}
Zhai, R.; Dan, C.; He, D.; Zhang, H.; Gong, B.; Ravikumar, P.; Hsieh, C.-J.;
  and Wang, L.
\newblock 2020.
\newblock Macer: Attack-free and scalable robust training via maximizing
  certified radius.
\newblock {\em arXiv preprint arXiv:2001.02378}.

\end{thebibliography}
\bibliographystyle{aaai}

\end{document}